\documentclass[10pt,twocolumn,letterpaper]{article}

\usepackage{graphicx}
\usepackage{amsmath}
\usepackage{amssymb}
\usepackage{booktabs}
\usepackage{multirow}
\usepackage{tabularray}
\usepackage{xcolor}

\definecolor{forestgreen}{rgb}{0.13, 0.55, 0.13}

\usepackage[pagebackref,breaklinks,colorlinks]{hyperref}

\usepackage[capitalize]{cleveref}
\crefname{section}{Sec.}{Secs.}
\Crefname{section}{Section}{Sections}
\Crefname{table}{Table}{Tables}
\crefname{table}{Tab.}{Tabs.}

\def\maketitlesupplementary
   {
   \newpage
       \twocolumn[
        {
        \centering
        \Large
        \textbf{MimicGait: A Model Agnostic approach for Occluded Gait Recognition using Correlational Knowledge Distillation}\\
        }
        {
        \large
        \centering
        \vspace{0.5em}Supplementary Material \\
        \vspace{1.0em}
        }
        
       ] 
   }

\begin{document}

\title{MimicGait: A Model Agnostic approach for Occluded Gait Recognition using Correlational Knowledge Distillation}

\author{Ayush Gupta\\
Johns Hopkins University\\
3400 N. Charles St\\
{\tt\small agupt120@jh.edu}
\and
Rama Chellappa\\
Johns Hopkins University\\
3400 N. Charles St\\
{\tt\small rchella4@jhu.edu}
}
\maketitle

\begin{abstract}
    Gait recognition is an important biometric technique over large distances. State-of-the-art gait recognition systems perform very well in controlled environments at close range. Recently, there has been an increased interest in gait recognition in the wild prompted by the collection of outdoor, more challenging datasets containing variations in terms of illumination, pitch angles, and distances. An important problem in these environments is that of occlusion, where the subject is partially blocked from camera view. 
    While important, this problem has received little attention. Thus, we propose \textbf{MimicGait}, a \textit{model-agnostic} approach for gait recognition in the presence of occlusions. We train the network using a \textit{multi-instance correlational distillation loss} to capture both inter-sequence and intra-sequence correlations in the occluded gait patterns of a subject, utilizing an auxiliary Visibility Estimation Network to guide the training of the proposed mimic network. We demonstrate the effectiveness of our approach on challenging real-world datasets like GREW, Gait3D and BRIAR.
    We release the code in 
    \href{https://github.com/Ayush-00/mimicgait}{https://github.com/Ayush-00/mimicgait}.
     \vspace{-2mm}
\end{abstract}    

\section{Introduction}
\label{sec:intro}
Gait is an important biometric feature which can be used for identifying humans \cite{park2021uniqueness}, especially when the face is not visible. There has been significant progress in the field of Gait Recognition - the problem of recognizing subjects based on their walking pattern. Gait recognition can be performed by placing wearable sensors on subjects \cite{gafurov2009sensors, gait-sensor-2}, however, such methods require the subjects' cooperation and are not scalable. With progress in computer vision, the popularity of gait recognition techniques using only vision-based modalities has risen significantly. As a result, gait has gained a unique importance among all biometric signatures as one of the few identifying characteristics in humans that can be captured effectively at a distance. 

\begin{figure}
\begin{center}
   \includegraphics[width=\linewidth]{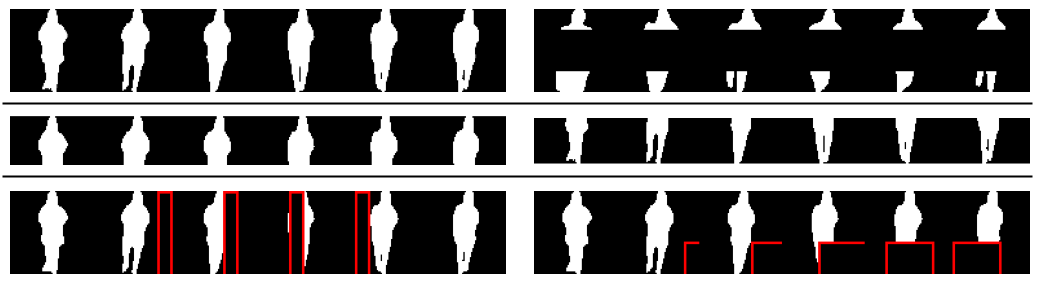}
\end{center}
\vspace{-3mm}
\caption{\label{fig:dynamic-consistent}Visualizations of the synthetic occlusions used in our experiments, taken from the GREW dataset. The original holistic video is shown in the top left. Middle occlusions are shown in the top right. The second row shows the same video with consistent synthetic occlusions. The bottom most row shows the same video with dynamic occlusions. The boundary of the moving occlusion patch is shown in red for visualization purposes only. 
}
\vspace{-1.5mm}
\end{figure}

There has been significant progress in the field of vision-based gait recognition \cite{shen2022comprehensive}, with some methods achieving almost perfect scores on indoor controlled datasets \cite{lin_gait_2021, opengait}. With saturation in controlled scenarios, there has been an increased focus on outdoor, in the wild scenarios \cite{zhu2021gait, Cornett_2023_WACV}. Such datasets pose much bigger challenges to gait recognition - due to large variations in viewpoint, altitude, clothing, background, illumination changes and occlusion. 

A deployed gait recognition system should be able to handle occlusion scenarios.
There can be many types of occlusions; arising from an obstruction between the camera and the subject, or due to improper camera placement. Occlusion can be consistent, such as an elevated sidewalk blocking a subject's feet for the entire sequence, or dynamic, such as another person or a stationary object temporarily blocking the subject of interest from view. With gait being recognized as a viable option for biometrics, it is important to address occlusion within gait recognition.

Most current work on gait recognition does not address this problem specifically; the lack of a large-scale dataset focused on occlusion has resulted in slow progress in the area. Some works that focus on this problem simulate occlusions on indoor datasets \cite{xu_occlusion-aware_2023} or work with small datasets \cite{singh2022hybrid}. Moreover, most current approaches assume an ideal occlusion scenario, where the subject is close to the camera and clearly visible. In a more practical in-the-wild scenario, the subject could be hundreds of meters away, and the camera may be situated at an altitude, and it is not easy to extend these approaches to such unconstrained data.

\cite{uddin2019spatio} uses a generative network to synthesize complete gait sequences in the case of occlusion. However, it will be difficult to generate such sequences when the partial input is itself of low quality. Similarly, \cite{xu_occlusion-aware_2023} uses an SMPL-based human mesh model to construct the gait signature, but the 3D structure of the body is not easy to recover from noisy data collected at a distance of several hundred metres, that too in the presence of occlusions. 
\cite{occ_aware} generates occlusion aware features and inserts them inside the gait recognition network. However, it is limited by the assumption that the network can independently learn discriminative features through an occlusion detector, neglecting the potential correlations between occluded and visible body parts.

To address these challenges,  we propose \textit{MimicGait}, a \textit{model-agnostic} approach to generate discriminative gait features for subjects at range under occlusion using correlational knowledge distillation. We assume that temporal patterns which exist in the occluded sections of the subject are correlated with the observable motion in the gait sequence. We adopt a knowledge distillation approach to learn these correlations among the occluded and visible parts of the body, enabling the network to produce features closer to the ideal, holistic features. 
Building upon the work by \cite{occ_aware}, we also utilize a Visibility Estimation Network (VEN) to introduce occlusion-relevant features into our method to enhance the prediction of these missing correlations.
Our approach is model-agnostic and can be used to increase the performance of any state-of-the-art gait recognition method to extend it to occlusions. Lastly, being a vision-based method, it can work with noisy data captured at a distance. We demonstrate the performance of our proposed method on the publicly available GREW \cite{zhu2021gait} and Gait3D \cite{gait3d} datasets. Additionally, we also evaluate our approach on the BRIAR \cite{Cornett_2023_WACV} dataset, which includes variations in range, altitude, clothing and walking conditions.

We also introduce some practical evaluation schemes for occluded gait recognition.
We formalize the concept of \textit{Generalizability}, introduced in \cite{occ_aware}, relating model performance on unseen occlusions.
We further introduce a concept of \textit{Adaptability}, relating to how well the model can be trained/adapted to newer occlusions. This becomes useful where a certain type of occlusion is expected to occur frequently in deployment.
Lastly, we propose another metric for occluded gait recognition called \textit{relative performance}, RP, which measures the occluded recognition performance of the model relative to performance on ideal, holistic data. 
We show that it is a better metric to evaluate occluded performance. Together with generalizability, adaptability and RP, we perform an extensive analysis of our method and show that it outperforms previous works. 

In summary, our main contributions are 

\begin{itemize}
    \item We propose \textbf{MimicGait}, a novel model-agnostic approach to generate robust gait features under occlusions in various conditions, viewpoints, and distances.
    
    \item We demonstrate the utility of a \textit{multi-instance correlational knowledge distillation} approach to learn the correlations between occluded and visible motion patterns across multiple gait instances of a subject. 
    
    \item We improve upon the auxiliary occlusion detector proposed in previous works and propose a \textit{Visibility Estimation Network} to enhance gait recognition performance under occlusions.

    \item We introduce the concepts of \textit{generalizability, adaptability,} and a new metric \textit{RP} for evaluating occluded gait recognition performance. Across these benchmarks, our approach outperforms other works on GREW, Gait3D, and the challenging real-world BRIAR dataset.

\end{itemize}

\begin{figure*}[ht]
\begin{center}
   \includegraphics[width=0.9\linewidth]{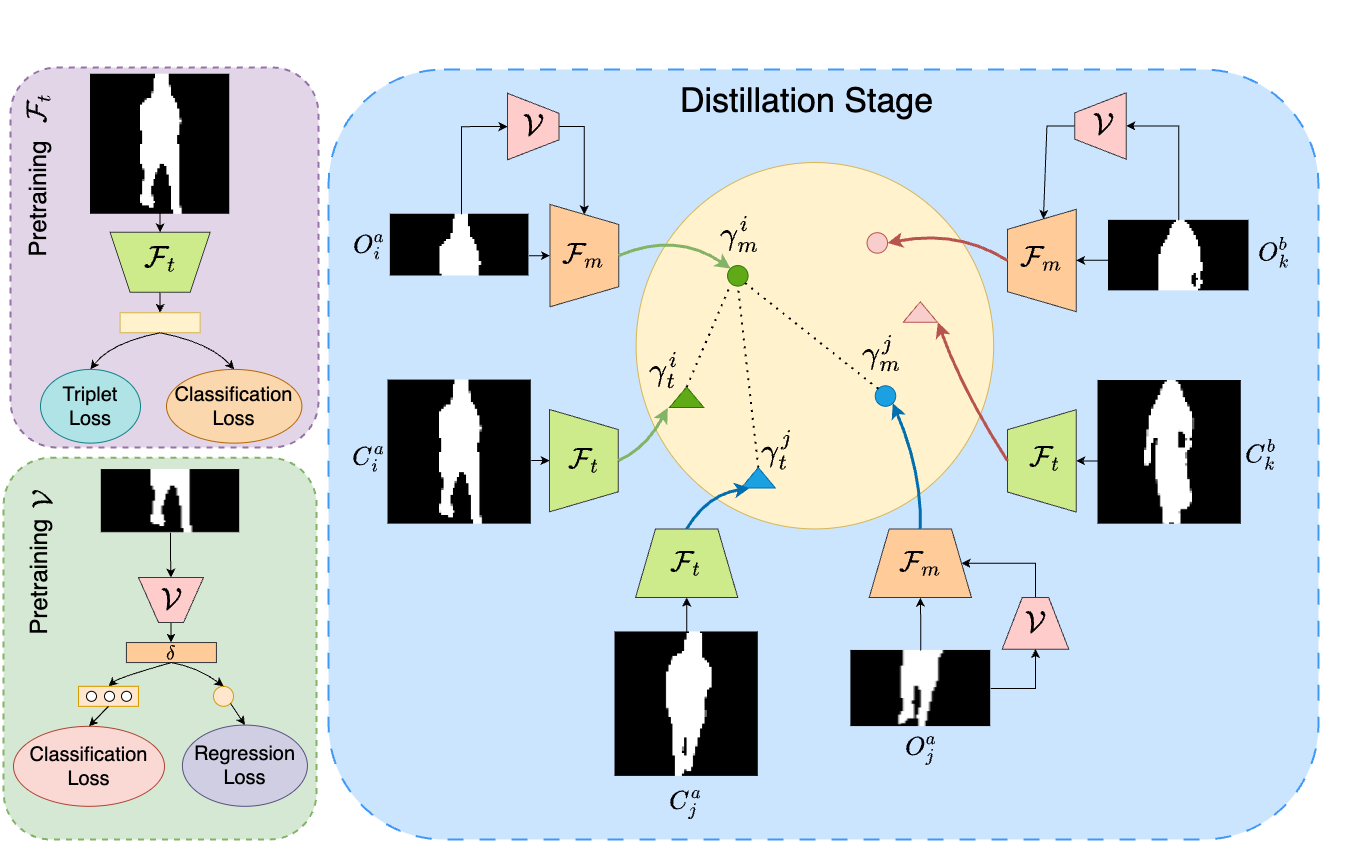}
\end{center}
\vspace{-2mm}
\caption{\label{fig:main}
Overview of the proposed approach. The training procedure consists of two stages. In the pretraining stage, the Visibility Estimation Network $\mathcal{V}$ (also called VEN) and the teacher network $\mathcal{F}_t$ are trained. In the distillation stage, a new mimic network $\mathcal{F}_m$ is trained by $\mathcal{F}_t$ using a multi-instance correlational KD loss. $\mathcal{V}$ is used to guide $\mathcal{F}_m$ by injecting occlusion-relevant features.
The $i^{th}$ video of subject $a$  may be occluded (denoted by $O_i^a$) or holistic (denoted by $C_i^a$). Their representations in the latent space are denoted by $\gamma_m$ and $\gamma_t$ respectively. The three types of anchor-positive pairs sampled by the proposed loss seen in the figure are described in \cref{sec:mickd-loss}.
}
\vspace{-4mm}
\end{figure*}

\vspace{-3mm}
\section{Related Work}
\label{sec:related_work}

\subsection{Gait Recognition}
Traditionally, gait recognition was performed using wearable motion sensors \cite{gafurov2009sensors, sensor-gait-survey}. With advances in computer vision techniques, gait recognition has become an attractive method for human identification at a distance \cite{deep-gait-survey}. These techniques can be classified into two categories based on the data modality; 1) Skeleton-based and 2) Vision-based. Skeleton-based gait recognition systems \cite{liao2017pose, lima2021pose, hasan2020pose, fan2023skeletongait, GPGait, GaitTR} first use pose estimation techniques \cite{pose-survey} to extract the joints/keypoints from the input image or video. This introduces a bottleneck in the form of the pose estimation method. Vision-based gait recognition systems usually operate on silhouettes \cite{zhu2021gait, lin_gait_2021, Fan_2020_CVPR, opengait, deepgaitv2}.
Some progress has been made to utilize the RGB modality  \cite{zhang2019gait} for gait recognition. However, RGB videos have a lot of irrelevant information like background, texture, and color. To account for this, works like \cite{Li_2020_ACCV, liang2022gaitedge} adopt an end-to-end approach while learning silhouettes. 

Performance of gait recognition methods has almost saturated \cite{lin_gait_2021, opengait} on indoor controlled datasets like CASIA-B \cite{yu2006framework}. To further advance research in this field, more challenging real-world datasets like GREW \cite{zhu2021gait}, Gait3D \cite{gait3d} and BRIAR \cite{Cornett_2023_WACV} have been collected with large variations in viewpoints, altitude, and other conditions. To overcome the new challenges posed by such outdoor datasets, works like \cite{yuxiang-multi-modal-gait} use multiple modalities for the recognition task.

\vspace{-1mm}
\subsection{Occluded Person Re-ID and Gait Recognition}
\vspace{-1mm}
Occlusions can severely hamper gait recognition and Person Re-ID systems\cite{reid-ex1, reid-survey}. Occlusion has already been recognized as an important problem in Person Re-ID \cite{occ-reid-survey}, with datasets targetting the occlusion problem specifically \cite{miao2019occ-reid-data, zhuo2018occ-reid-data}. Current works deal with simulated occlusions \cite{chen_occlude_2021} as well as real occlusions \cite{miao_identifying_2022}. \cite{chen_occlude_2021} categorizes occlusions into different broad categories and uses this knowledge to better identify the subject, while \cite{miao_identifying_2022} uses pose estimation techniques to identify visible body parts, generating different representations for each part. 

Unlike the Person Re-ID problem, occlusion has not received much attention in gait recognition due to lack of large-scale datasets. Some available works focus on reconstruction \cite{singh2022hybrid, uddin2019spatio} of the silhouette sequence using generative networks or estimate an SMPL-based 3D mesh model \cite{xu_occlusion-aware_2023} to infer the missing body parts. 
However, these approaches are not easily extendable to long ranges or noisy data, as would be the case in a real-world scenario.

Some works simulate occlusions in existing datasets and utilize auxiliary networks to gain additional information about occlusions \cite{Xu_2023_ICCV, occ_aware}. \cite{Xu_2023_ICCV} performs silhouette registration to enhance the input, but only works with top occlusions on indoor datasets. \cite{occ_aware} works on outdoor data and a diverse set of occlusions, but neglects the potential correlations between occluded and visible body parts due to exclusive exposure to occluded data during training.

\vspace{-1mm}

\subsection{Knowledge Distillation}
\vspace{-1mm}
\label{sec:related-work-kd}
Broadly, Knowledge Distillation (KD) involves a student-teacher learning framework where one model passes on its `knowledge' to another model \cite{kd-survey}. Initially, KD was used as a technique for model compression and acceleration \cite{hong2022compression-kd}, where a large model was used to train a smaller network to reduce memory and computing requirements. However, the utility of KD was shown in other areas as well. \cite{radosavovic2018data-distil} used the idea of data distillation to train a student network using a single teacher by applying multiple transforms on the input. \cite{hong2020dehaze-distil} used distillation to teach a student network to dehaze an image by applying consistency loss in intermediate features between the student and teacher. \cite{peng2019correlation} proposes correlational congruence in knowledge distillation, utilizing correlation across multiple instances to teach their student network. \cite{kd_gait} uses KD in the context of gait recognition, transferring knowledge from RGB to silhouette encoders to infer 3D body features.

To work with occluded data captured from a distance, we utilize KD to capture correlations among the occluded and visible body parts to learn a gait signature for the subject.   
\section{Proposed Method}

Given an input video sequence $S^{i} = \{v_{1}, v_{2}, ... v_{n}\}$ for subject $i$, the goal is to find a discriminative gait signature $\gamma$ for the subject. We assume that the input sequence consists of binary silhouette masks. In the occlusion scenario, this mask may not be completely visible in some or all of the frames which makes the task more challenging. 

The overview of the proposed approach is shown in \cref{fig:main}. The entire training procedure consists of two stages - the pretraining stage, and the distillation stage. In the pretraining stage, a state-of-the-art gait recognition model $\mathcal{F}_t$ is trained on the original unoccluded video frames. Separate from $\mathcal{F}_t$, the VEN $\mathcal{V}$ is also trained to identify the amount and type of synthetic occlusions present in the video. These two networks are used in the next stage with frozen weights.

In the distillation stage, a new network $\mathcal{F}_m$, the mimic network, is initialized with the same architecture as $\mathcal{F}_t$. $\mathcal{F}_m$ is trained to output a gait signature $\gamma_m$ by taking the occluded video $O^{i}$ as input.
At the same time, $\mathcal{F}_t$ outputs the gait signature $\gamma_t$ by taking the corresponding full body video $C^{i}$ as input.  A multi-instance correlational distillation loss is used to train $\mathcal{F}_m$ to bring the distributions of $\gamma_m$ and $\gamma_t$ closer in the latent space. 
VEN is used to regulate $\mathcal{F}_m$ by injecting visibility information into the backbone.
The method is described in detail in the following sections.

\subsection{Visibility Estimation Network}
\label{sec:visibility-estimation-description}
VEN is a CNN that predicts the type of occlusion, if any, present in the input video and also a measure of the amount of this occlusion, which we call visibility estimation. 
This module is inspired by \cite{occ_aware}, where an auxiliary network was used to identify the occlusion class. VEN builds on top of \cite{occ_aware} by performing the visibility estimation task while simultaneously predicting the occlusion type.

In the pretraining stage, VEN is trained on synthetic occlusions. Some examples of the occlusions are shown in Figure \ref{fig:dynamic-consistent}. It consists of a sequence of convolutional and linear layers, with two parallel heads - one classification head for the occlusion classification task, and one regression head for the visibility estimation task. Accordingly, cross entropy and L2 regression losses are used on the two heads to train the network, making it learn occlusion-relevant features.  

When VEN is used in the distillation stage, the two heads are removed and the penultimate feature vector $\delta$ is used to guide $\mathcal{F}_m$. It is important to note that the weights of VEN are frozen during this stage, to ensure that it retains the visibility awareness learned during training.

\subsection{Mimic Network}
\vspace{-1mm}
\label{sec:mimic-description}
The main idea behind the mimic network is that the features corresponding to the missing/occluded body parts are correlated with the observable motion of the subject. Every moving body part is correlated to a global pattern and also has its own distinctive local motion, which is why pyramid structures operating on multi-scale input are popular in gait recognition \cite{lin_gait_2021, Fan_2020_CVPR}. It is these patterns of motion that constitute gait, and when some of these patterns are missing, the mimic network uses its learned correlations and the available input to fill in the gaps in the gait signature.

During the pretraining stage, a state-of-the-art gait recognition backbone $\mathcal{F}_t$ is trained on the original, unoccluded videos to generate discriminative gait signatures $\gamma_t$. 
During the distillation stage, $\mathcal{F}_m$ is trained to output a discriminative gait signature $\gamma_m$ using the occluded videos $O^i$ as input. Since occlusions can be of many different types, and the internal architecture of $\mathcal{F}_m$ does not target any occlusion, we use the occlusion-relevant features from VEN to guide $\mathcal{F}_m$, similar to \cite{occ_aware}.
Specifically, 
\begin{equation}
    \gamma_m = \mathcal{F^{'}}_m(O^i) = T(\mathcal{F}_m(O^i) \oplus \mathcal{V}(O^i))
\end{equation}
where $\mathcal{V}$ refers to VEN, $\oplus$ is the concatenation operation, and $T$ is a linear transformation to make the feature size compatible after concatenation. By introducing visibility features from VEN into $\mathcal{F}_m$ in such a manner, the network gains information about the visibility of the subject in the input, which in turn helps it to generate features closer to the optimal holistic features.  

In the distillation stage, $\mathcal{F}_t$ acts as a teacher network and $\mathcal{F}^{'}_m$ tries to mimic the holistic features generated by $\mathcal{F}_t$. However, the difference is that $\mathcal{F}^{'}_m$ is only allowed to see the occluded videos during training. 
In the process of trying to mimic the holistic features from the occluded inputs, $\mathcal{F}_m$ is able to learn better gait representations for occlusion scenarios. This is achieved by using a multi-instance correlational distillation loss as described in the next section. 

\subsection{Multi-instance Correlational Distillation Loss}
\label{sec:mickd-loss}
Inspired by \cite{peng2019correlation}, we use a multi-instance correlational KD loss modified for the occluded gait recognition task. We recognize that while there is correlation within the motion of different body parts of the same gait instance, there is also correlation among the motion of the body parts across multiple gait instances of the same subject. The distillation loss, modeled as a Triplet Loss \cite{balntas2016triplet}, is able to capture both these correlations. As shown in \cref{fig:main}, the outputs of the mimic network, $\gamma_m$, and those of the teacher network, $\gamma_t$, are used to sample anchor-positive pairs of the following three types: 1) $\gamma_m^i$ - $\gamma_t^i$, representing student-teacher learning within the same gait instance, 2) $\gamma_m^i$ - $\gamma_t^j$, representing student-teacher learning across instances, and 3) $\gamma_m^i$ - $\gamma_m^j$, representing mimic network correlation across instances. 

These considerations lead to a triplet margin loss:
\vspace{-1mm}
\begin{equation}
    L = \sum_{i}[ D^i_{a,p} - D^i_{a,n} + m ]_+
\vspace{-2mm}
\end{equation}

where the summation is over all anchor-positive-negative triplets and $D_{a,p}$, $D_{a,n}$ refer to the Euclidean distance between the anchor-positive (AP) or anchor-negative (AN) pairs. The AP pairs are sampled from the previously mentioned three types, while the AN pairs are sampled from other identities, and $m$ is the margin.

\textbf{Inference: }
During inference, when only the occluded video is available, the mimic network $\mathcal{F}_m$ guided by the VEN is used to generate the gait features.
\section{Experimental Setup}

\subsection{Datasets}
\begin{figure}
\vspace{-5mm}
\begin{center}
   \includegraphics[width=0.95\linewidth]{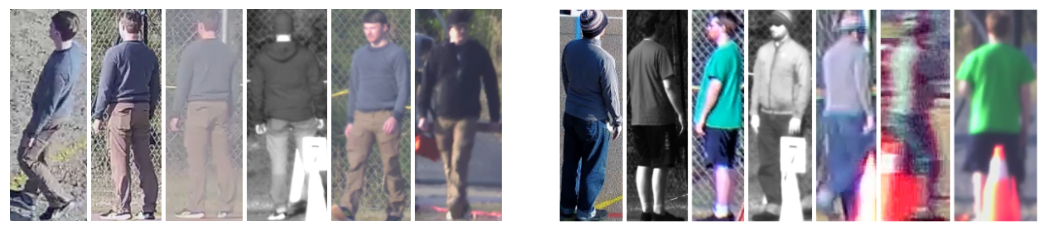}
\end{center}
\vspace{-5mm}
\caption{\label{fig:briar-examples}Some samples images taken from the BRIAR dataset for two subjects. From left to right, the range of capture increases from close range to 1000m for each subject. A large variation in the quality of the videos and the collection conditions can be seen. Subjects have consented to the use of these images in publication.
}
\vspace{-4mm}
\end{figure}

\paragraph{\textbf{BRIAR:}}
We use the BRIAR \cite{Cornett_2023_WACV} dataset to conduct our experiments. This is a recently collected dataset that contains many variations in the walking conditions of subjects. The subset of BRIAR data we use comprises of 776 training subjects and 856 test subjects.
The dataset contains videos of subjects walking in indoor, controlled environments as well as outdoor field environments. The outdoor data is captured from many different camera sensors, ranges, altitudes and viewpoints. Some example images are shown in \cref{fig:briar-examples}.

Videos of walking subjects in outdoor environments are captured systematically at distances ranging from 100m to 1000m. Additionally, videos are also captured using an UAV and at close range with extreme viewpoint. 

Each subject either walks randomly inside a fixed boundary (random), or along well-defined straight lines (structured), or both, in different videos. The subject may be carrying a large object like a big cardboard box in some videos, while some may be using their cellphones while walking.

\vspace{-4mm}

\paragraph{\textbf{GREW:} }

GREW \cite{zhu2021gait} is a large scale publicly available dataset for gait recognition, comprising of 20,000 subjects in the training split and 6,000 in the testing split. The dataset is captured from in-the-wild videos in a variety of conditions from multiple different cameras with varying viewpoints. We utilize the provided 2D silhouettes in this work. 

\vspace{-3mm}

\paragraph{\textbf{Gait3D:} }
Gait3D \cite{gait3d} is a publicly available in-the-wild gait recognition dataset, comprising of 3,000 subjects in the training set and 1,000 subjects in the testing set. We utilize the provided 2D silhouettes of the dataset in this work. 


\subsection{Synthetic Occlusions}
\label{sec:synthetic-occ-desc}
Broadly, occlusions can be classified into two categories - 1) consistent, where the occlusion stays roughly the same for the entire length of the video, and 2) dynamic, where they can change with time.


In uncontrolled environments, consistent occlusions occur when there are obstacles like an elevated sidewalk or tall grass between the subject and the camera, or they might arise due to a bad camera angle. Dynamic occlusions occur when objects, or other people, temporarily block the subject of interest from view.
We try to simulate both consistent and dynamic occlusions by placing stationary or moving black patches on the input frames, as shown in \cref{fig:dynamic-consistent}. For consistent occlusions, we remove either the top, bottom or middle part of the frame.
Other works on occluded gait recognition\cite{occ_aware} simulate more type of occlusions, however many of these occlusions are not likely to occur in a practical scenario. We focus on the the top and bottom occlusions in our main results and perform generalizability and adaptability evaluations using middle and dynamic occlusions in \cref{sec:experiments}.

Occlusions are introduced randomly in the videos during training and evaluation. We randomly occlude a portion of the frame as shown in \cref{fig:dynamic-consistent}. The amount of occlusion is selected randomly within a fixed range $R$, which we set to (0.4, 0.6) for all our experiments. This means that the portion of the synthetic occlusion is chosen randomly between 40\%-60\% of the spatial dimensions of the input video. 
More details about the synthetic occlusions have been included in the supplementary material.

\begin{table*}
\centering
\resizebox{\linewidth}{!}{%
\begin{tabular}{c|c||cc||cc||ccc}
\multirow{2}{*}{\textbf{Backbone}}   & \multirow{2}{*}{\textbf{Method}} & \multicolumn{2}{c||}{\textbf{Gait3D}}         & \multicolumn{2}{c||}{\textbf{GREW}}           & \multicolumn{3}{c}{\textbf{BRIAR}}                                     \\ 
\cline{3-9}
                                     &                                  & \textbf{Rank-1}       & \textbf{Rank-5}       & \textbf{Rank-1}       & \textbf{Rank-5}       & \textbf{Rank-1}       & \textbf{Rank-20}      & \textbf{TAR@0.01 FAR}  \\ 
\hline
\multirow{4}{*}{\textbf{GaitBase}}   & Baseline-1                       & 7.6 (0.11)            & 15.71 (0.19)          & 14.85 (0.27)          & 25.55 (0.35)          & 1.34 (0.04)           & 12.05 (0.16)          & 2.46 (0.04)            \\
                                     & Baseline-2                       & 17.12 (0.24)          & 31.43 (0.37)          & 16.42 (0.30)          & 30.38 (0.42)          & 6.13 (0.16)           & 27.52 (0.36)          & 9.81 (0.17)            \\
                                     & Occlusion Aware \cite{occ_aware}                  & 18.22 (0.26)          & 34.94 (0.41)          & 22.55 (0.41)          & 37.95 (0.53)          & 8.70 (0.23)           & 35.38 (0.46)          & 12.83 (0.22)           \\
                                     & \textbf{Mimic Network (ours)}    & \textbf{22.72 (0.32)} & \textbf{40.84 (0.48)} & \textbf{28.38 (0.51)} & \textbf{45.43 (0.63)} & \textbf{10.93 (0.28)} & \textbf{42.25 (0.55)} & \textbf{12.92 (0.22)}  \\ 
\hline
\multirow{4}{*}{\textbf{GaitGL}}     & Baseline-1                       & 3.00 (0.11)           & 6.70 (0.15)           & 3.60 (0.10)           & 6.90 (0.13)           & 0.69 (0.05)           & 6.70 (0.18)           & 1.90 (0.09)            \\
                                     & Baseline-2                       & 4.4 (0.16)            & 9.71 (0.22)           & 6.3 (0.17)            & 11.92 (0.22)          & 1.94 (0.15)           & 13.17 (0.35)          & 3.42 (0.16)            \\
                                     & Occlusion Aware \cite{occ_aware}                  & 4.8 (0.18)            & 12.7 (0.29)           & 7.05 (0.19)           & 12.72 (0.24)          & 4.02 (0.30)           & 22.85 (0.61)          & 4.19 (0.19)            \\
                                     & \textbf{Mimic Network (ours)}    & \textbf{5.6 (0.21)}   & \textbf{13.5 (0.31)}  & \textbf{7.53 (0.21)}  & \textbf{13.97 (0.26)} & \textbf{5.4 (0.40)}   & \textbf{24.75 (0.66)} & \textbf{5.14 (0.23)}   \\ 
\hline
\multirow{4}{*}{\textbf{DeepGaitV2}} & Baseline-1                       & 2.20 (0.03)           & 6.21 (0.07)           & 3.72 (0.05)           & 6.85 (0.08)           & 2.38 (0.05)           & 11.40 (0.13)          & 1.80 (0.03)            \\
                                     & Baseline-2                       & 6.71 (0.09)           & 14.21 (0.16)          & 9.65 (0.13)           & 16.20 (0.19)          & 6.52 (0.13)           & 27.47 (0.32)          & 7.40 (0.11)            \\
                                     & Occlusion Aware \cite{occ_aware}                  & 14.01 (0.18)          & 27.83 (0.32)          & 13.38 (0.18)          & 22.87 (0.27)          & 7.39 (0.15)           & 32.74 (0.38)          & 7.12 (0.10)             \\
                                     & \textbf{Mimic Network (ours)}    & \textbf{16.82 (0.22)} & \textbf{33.03 (0.38)} & \textbf{14.38 (0.20)} & \textbf{24.93 (0.29)} & \textbf{11.49 (0.24)} & \textbf{44.71 (0.52)} & \textbf{20.73 (0.30)}  
\end{tabular}
}
\vspace{-2mm}
\caption{\label{tab:main-results-mimic}
Our results on the Gait3D \cite{gait3d}, GREW \cite{zhu2021gait} and BRIAR \cite{Cornett_2023_WACV} datasets, for different gait recognition backbones. Baseline-1 refers to zero shot evaluation on occluded data. Baseline-2 refers to training the network on occlusions. The values in (.) denote the relative performance (RP) values with respect to the ideal, no occlusion scenario. We can see that the mimic network outperforms other methods on occluded data, achieving at least 20-30\% RP across datasets and backbones.
}
\vspace{-5mm}
\end{table*}

\subsection{Baselines}
To showcase the model-agnosticity of our method, we experiment with existing architectures like GaitBase \cite{opengait}, GaitGL \cite{lin_gait_2021} and deeper networks like DeepGaitV2 \cite{deepgaitv2}.

Following \cite{occ_aware}, we train these networks on holistic videos and evaluate them directly on synthetic occlusions in a zero-shot setting. This is called Baseline-1. 
Since these architectures do not address the occlusion problem specifically, we retrain them on synthetically occluded data as Baseline-2. We also compare our results with \cite{occ_aware}.

\subsection{Implementation Details}
\paragraph{\textbf{Visibility Estimation Network:} }
VEN consists of three convolutional layers and two linear layers. VEN consists of two heads - a classification head for classifying occlusions and a regression head for visibility estimation.
In our experiments, VEN is trained to classify the input into three classes, namely, no occlusion, top occlusion and bottom occlusion. This set is updated as more occlusion types are introduced for training the mimic network. More details about VEN are included in the supplementary material.

\vspace{-2mm}
\paragraph{\textbf{Mimic Network:}}
We localize the subjects in an $H\times W = 64\times 64$ bounding box during preprocessing and crop these bounding boxes from the video. The mimic network architecture is the same as the gait recognition backbone $\mathcal{F}_t$. Training occurs in two stages. We train $\mathcal{F}_t$ on holistic videos in the pretraining stage. Next, we train $\mathcal{F}_m$ using features obtained by $\mathcal{F}_t$ in the distillation stage. We choose the same optimizer as used by $\mathcal{F}_t$ during pretraining, namely Adam \cite{KingmaB14adam} for GaitGL and SGD for GaitBase and DeepGaitV2. The learning rate for GaitGL experiments is 1e-4, while for GaitBase and DeepGaitV2 it is 1e-1. Additional details about the training procedure of the mimic network have been included in the supplementary material.

\subsection{Evaluation Metrics}

\vspace{-1mm}
\paragraph{\textbf{Rank Retrieval:} }
Rank retrieval accuracy is a standard metric to evaluate recognition performance. 
We follow the gallery-probe splits of \cite{opengait} for GREW and Gait3D.
For GREW, the probe set labels are not publicly released. To enable local evaluation for the GREW dataset, we follow the method proposed in \cite{opengait}, the details of which are provided in the supplementary material. 
The BRIAR dataset provides its own protocol \cite{Cornett_2023_WACV}. We additionally compute the verification performance for BRIAR.

\textbf{Relative Performance (RP): }
In the model-agnostic scenario being discussed in this work, we need to measure the effectiveness of an \textit{occlusion-mitigating method} across backbones. Considering only the performance of a model on occluded data - the \textit{occluded performance} $OP$ - gives an incomplete picture. 
This is because a low $OP$ might be caused by other factors not related to the strength of the occlusion-mitigating method - such as the backbone being suboptimal or the dataset being too difficult.
To filter out these other factors and focus on just the strength of the occlusion-mitigating method, we define a new RP metric,

\begin{equation}
    RP = \frac{OP}{HP}
\end{equation}

where $OP$ is the occluded performance and $HP$ is the holistic performance on unoccluded data for a given backbone.
If the backbone is suboptimal or the dataset is too difficult, both $OP$ and $HP$ are low, therefore the RP is not affected very much. 
However, if the occlusion-mitigating method is suboptimal, only the $OP$ is low - reducing the RP. 
This makes RP relatively more suited to model-agnostic evaluation of occlusion-mitigating methods, like the one introduced by \cite{occ_aware} and our proposed mimic network.

Another way of looking at RP is that it normalizes the $OP$ by its true upper bound $HP$, so changes in $OP$ are measured with respect to $HP$.
A small improvement $\Delta y$ in $OP$ may seem insignificant, but becomes important if $HP$ itself is small - for example, an improvement of 1\% in Rank-1 accuracy in $OP$ is significant if the upper bound $HP$ is itself just 10\%! 

\begin{figure}

\begin{center}
   \includegraphics[width=0.8\linewidth]{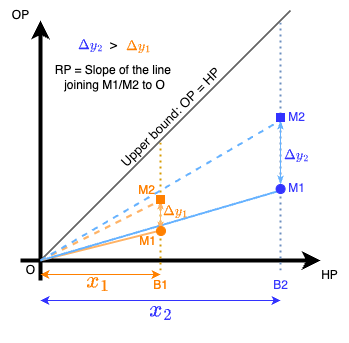}
\end{center}
\vspace{-9mm}
\caption{\label{fig:rp-visualization}
Comparing two hypothetical occlusion-mitigating methods M1/M2 between two backbones B1/B2 on occluded performance ($OP$) and  holistic performance ($HP$).
A small change $\Delta y_1$ in $OP$ can cause a large change in the slope/RP for B1, but a larger $\Delta y_2$ is needed to cause a similar change in slope for B2.
By considering the slope rather than just the $OP$, RP is able to better isolate the effect of M1/M2 across backbones.
}
\vspace{-4mm}
\end{figure}

\cref{fig:rp-visualization} provides a geometric interpretation of this scenario.
B1/B2 are two hypothetical backbones and M1/M2 are two occlusion-mitigating methods. RP is the slope of the line joining origin to M1/M2.
For the suboptimal backbone B1, a small $\Delta y_1$ by using M2 over M1 can cause a large change in slope/RP. 
On the other hand, a larger increase of $\Delta y_2$ in $OP$ is needed to cause similar improvements in slope/RP.
Thus, RP gives more insight into the improvement brought about by M2 in this backbone-agnostic scenario by normalizing $OP$ by the performance of the backbone.

\textbf{Adaptability:}
This test aims to evaluate how well a model adapts to new occlusion types when it is further trained on them along with the original occlusions. 
This test provides a new perspective - it analyses the scenario when we want to extend the model's capability to a new occlusion type by additional training.
The results of the adaptability evaluation are presented in \cref{tab:new-occs}.

\textbf{Generalizability: }
While having prior knowledge about occlusion types is an advantage, it is not practical to train on all possible occlusion types. 
For a method to be deployable, it should be able to generalize to occlusions not seen during training. 
This idea of generalizability to different occlusions was introduced in \cite{occ_aware}. In the generalizability test, we take models trained on top and bottom occlusions and evaluate them directly on the newer occlusion types, in a zero-shot setting.
The results are presented in \cref{tab:new-occs}.

\section{Results and Discussion}

\label{sec:experiments}

In general, we observe that the mimic network outperforms other methods for occluded gait recognition. This statement holds across datasets and backbones, demonstrating the effectiveness and model-agnosticity of our method (\cref{tab:main-results-mimic}). The generalizability and adaptability results presented in \cref{tab:new-occs} confirm that the mimic network also outperforms other approaches here. Thus, we conclude that capturing correlations among occluded and visible body parts using our proposed mimic network does indeed help in occluded scenarios. However, apart from this general trend, we make several interesting observations. 


\paragraph{Insights from the RP metric: }
In some cases, performance improvements are not visible just from Rank-K accuracy, such as with GaitBase on BRIAR or GaitGL on Gait3D, due to low absolute performance. 
This is attributed to the challenging nature of the datasets (e.g., BRIAR with extreme range and turbulence) or lower capability of backbones (e.g., GaitGL). 
The RP metric can filter out these factors to some extent. It is not perfect at filtering these factors, and RP might still change across backbones, but improvements are easier to see in RP even in the case of suboptimal backbones or difficult datasets.


\paragraph{Increasing RP with Rank:}
An interesting observation is the effect of rank. As rank increases, both absolute accuracy and RP increase across all experiments. 
The increase in RP is non-trivial because RP is the ratio of occluded to holistic accuracy, and the rising RP indicates that occlusion accuracy grows faster than holistic accuracy as rank increases! 
This is possibly because the model is better at leveraging partial information to correctly identify occluded instances when given more opportunities (higher rank) to match. Holistic data already benefits from complete information, leading to a relatively slower increase in accuracy.


\paragraph{Deeper networks:} DeepGaitV2 is considered to be better than GaitBase as a backbone \cite{deepgaitv2}. Interestingly, we observe that it performs worse under occlusions in many of our experiments. 
We think that the larger number of parameters and the larger depth of DeepGaitV2 make it harder to optimize under occluded conditions where the input is more sparse. 
Regardless, the mimic network still performs better than the occlusion aware method for this backbone as well.


\paragraph{Relative difficulty of different occlusions:}
Comparing \cref{tab:main-results-mimic} and the adaptability section \cref{tab:new-occs} suggests that middle and dynamic occlusions are easier than top and bottom occlusions, since the model is able to perform better on the former set. We investigate this further  in the supplementary material by evaluating on these occlusions individually.

\begin{table*}[]
\centering
\resizebox{0.9\linewidth}{!}{%
\begin{tabular}{c||c|cc|cc}
\multirow{2}{*}{\begin{tabular}[c]{@{}c@{}}Additional \\Occlusion Type\end{tabular}} & \multirow{2}{*}{Method}       & \multicolumn{2}{c|}{Generalizability}         & \multicolumn{2}{c}{Adaptability}               \\
                                                                                     &                               & Rank1                 & Rank5                 & Rank1                 & Rank5                  \\ 
\hline
\multirow{3}{*}{Middle}                                                              & Baseline-1                    & 13.12 (0.24)          & 24.62 (0.34)          & -          & -            \\
                                                                                     & Baseline-2                    & 13.87 (0.25)          & 25.72 (0.36)          & 21.25 (0.38)          & 36.5 (0.51)            \\
                                                                                     & Occlusion Aware \cite{occ_aware}             & 17.93 (0.32)          & 32.15 (0.45)          & 26.7 (0.48)           & 43.82 (0.61)           \\
                                                                                     & \textbf{Mimic Network (ours)} & \textbf{21.73 (0.39)} & \textbf{37.37 (0.52)} & \textbf{34.78 (0.63)} & \textbf{52.75 (0.73)}  \\ 
\hline
\multirow{3}{*}{Dynamic}                                                             & Baseline-1                    & 17.27 (0.31)          & 28.05 (0.39)          & -          & -            \\
                                                                                     & Baseline-2                    & 17.48 (0.32)          & 31.53 (0.44)          & 32.1 (0.58)           & 49.68 (0.69)           \\
                                                                                     & Occlusion Aware \cite{occ_aware}               & 21.27 (0.38)          & 36.5 (0.51)           & 34.87 (0.63)          & 52.07 (0.72)           \\
                                                                                     & \textbf{Mimic Network (ours)} & \textbf{26.77 (0.48)} & \textbf{42.9 (0.60)}  & \textbf{36.65 (0.66)} & \textbf{53.15 (0.74)} 
\end{tabular}
}
\caption{\label{tab:new-occs}
The Adaptability (additional training) and Generalizability (zero-shot) evaluations using the GaitBase backbone on GREW dataset, using additional occlusion types. RP values are shown in (.). Note that adaptability is not applicable for Baseline-1, since Baseline-1 is not trained on any occlusions.
The mimic network outperforms other approaches in these scenarios. 
}
\end{table*}

\subsection{Ablation Studies}

In this section, we perform various ablations on the proposed network, removing or modifying different parts of the model to see their effects on performance. For all the experiments in this section, we use the GREW \cite{zhu2021gait} dataset and the GaitBase \cite{opengait} backbone unless stated otherwise.


\textbf{Effect of Multi-instance Correlational KD loss:}

In this section, we analyse the effect of the proposed multi-instance correlational distillation (MiCKD) loss on the network. 
To isolate the effects of the MiCKD loss, we remove the VEN in the experiments in this section, and deal with a `Vanilla Mimic' network. 
If we completely remove the distillation loss from this Vanilla Mimic network, the method becomes the same as Baseline-2, which does not capture any occluded-visible body part correlations. 
Next, we try a simpler approach for the distillation stage by considering correlations among $\gamma_m$ and $\gamma_t$ only within the same instance, minimizing the L2 distance between them (L2 KD). 
Comparing this to MiCKD in \cref{tab:ablation-kd} isolates the effect of utilizing multiple instances for feature learning.

We observe that the latter performs better, possibly because local gait patterns remain consistent across walking instances. The model is able to leverage this consistency to learn correlations among body parts which are occluded in one instance but visible in another.

\textbf{Adding Cross Entropy Loss: } Based on the training techniques of $\mathcal{F}_t$, we hypothesize that adding cross entropy loss using a BNNeck layer as done in \cite{opengait} would further help the model along with MiCKD loss.
However, as shown in \cref{tab:ablation-kd}, our hypothesis is negated and we observe that this approach actually reduces the model performance. We are unsure why this occurs, and hypothesize that the losses could conflict with each other.
Based on this, we choose to exclude cross-entropy loss in the final model.

\begin{table}
\centering
\resizebox{0.8\linewidth}{!}{
\begin{tabular}{c|cc}
Method                                   & Rank-1         & Rank-5         \\ 
\hline
No KD (Baseline-2)                       & 16.42          & 30.38          \\
L2 KD                                    & 20.43          & 35.95          \\
\textbf{MiCKD (Vanilla Mimic)} & \textbf{25.75} & \textbf{42.2}  \\
MiCKD + XE     & 20.22          & 34.93         
\end{tabular}
}
\caption{\label{tab:ablation-kd}Different distillation strategies for the mimic network, using GaitBase on the GREW dataset. We see that the proposed multi-instance correlational knowledge distillation loss (MiCKD) indeed helps in learning better occlusion features.}
\end{table}

\begin{table}
\centering
\resizebox{0.95\linewidth}{!}{%
\begin{tabular}{c|c|cc|cc}
 & \multirow{2}{*}{Mimic} & \multicolumn{2}{c|}{Proxy Tasks} & \multicolumn{2}{c}{Accuracy}     \\
     Method                   &                        & Classif.           & Reg.                 & Rank-1         & Rank-5          \\ 
\hline
Baseline-2              &                        &             &                    & 16.42          & 30.38           \\
Occlusion aware\cite{occ_aware}         &                        & \checkmark          &                    & 22.55          & 37.95           \\
VEN                     &                        & \checkmark          & \checkmark                 & 23.52          & 39.68           \\ 
\hline
Vanilla Mimic           & \checkmark                     &             &                    & 25.75          & 42.2            \\
Mimic + Occlusion Aware & \checkmark                     & \checkmark          &                    & 27.52          & 44.15           \\
\textbf{Mimic + VEN}    & \textbf{\checkmark}            & \textbf{\checkmark} & \textbf{\checkmark}        & \textbf{28.38} & \textbf{45.43} 
\end{tabular}
}
\caption{\label{tab:ablation-mimic-heads}
Effect of the mimic training strategy and the proxy tasks involved in the pretraining of the occlusion detector. Classif. and Reg. refer to the classification and regression tasks respectively. In general, training the auxiliary network on more tasks and using the mimic training strategy improves performance under occlusions.
}
\end{table}

\textbf{Effect of guidance by VEN: } 

In this section, we analyze the role of VEN in the mimic network. We compare the vanilla mimic network with the `Mimic + VEN' row of \cref{tab:ablation-mimic-heads}. The mimic network benefits from VEN guidance; without VEN, the network must independently determine which body parts are visible, complicating the extraction of gait patterns. In contrast, external guidance from VEN simplifies this task, allowing the network to focus on gait pattern extraction rather than occlusion identification.

\textbf{Different proxy tasks for training VEN: }
In the VEN pretraining stage, we train it to jointly output the occlusion type and the occlusion amount, with two different loss functions for each task. In this section, we investigate how useful these two individual tasks are for learning useful occlusion aware features. As such, we train a network without the occlusion amount regression head, making it similar to the occlusion detector in \cite{occ_aware} and compare it with VEN.

To get an estimate of the quality of occlusion awareness within these two networks, we compare the performance of gait recognition backbones trained using VEN and the occlusion detector in \cref{tab:ablation-mimic-heads}. We observe that VEN has inherently better occlusion relevant features, regardless of whether the mimic training strategy is applied or not - and so we conclude that the classification and regression proxy tasks individually contribute to performance.

\section{Limitations and Future Work}

Although our proposed method can perform better on synthetic occlusions, it is not perfect.
We proposed a general approach without altering the backbone,
and future works can explore  incorporating specific architectural changes to address occlusions better.
Further, we could not test our approach on real occlusions due to the absence of an occlusion category in the outdoor datasets we used. To properly evaluate our method, a large-scale dataset specifically focused on occlusions is essential to advance research in this area.
Lastly, we were unable to explore why adding cross entropy loss hurts MimicGait. Future work can explore this further to achieve more gains in performance.

\vspace{-2mm}

\section{Conclusion}
In this work, we proposed \textit{MimicGait}, a novel model-agnostic approach for occluded gait recognition. We proposed a multi-instance correlational KD loss to train the mimic network in a student-teacher setting, utilizing an auxiliary Visibility Estimation Network to introduce occlusion-relevant features.
We introduced generalizability and adaptability tests along with a new metric RP to evaluate occluded performance.
We evaluated our approach on GREW, Gait3D and BRIAR datasets, and showed that the proposed mimic network outperforms existing approaches on occlusions on real-world data collected from large distances.

\paragraph{Acknowledgements: }
This research is based upon work supported in part by the Office of the Director of National Intelligence (ODNI),
Intelligence Advanced Research Projects Activity (IARPA), via [2022-21102100005]. The views and conclusions contained herein are those of the authors and should not be interpreted as necessarily representing the official policies, either expressed or implied, of ODNI, IARPA, or the U. S. Government. The US. Government is authorized to reproduce and distribute reprints for governmental purposes notwithstanding any copyright annotation therein.

\newpage

{\small
\bibliographystyle{ieee_fullname}
\bibliography{egbib}
}

\clearpage
\setcounter{page}{1}

\maketitlesupplementary

\section{Introduction}

In this supplementary material section, we first provide more details about the BRIAR dataset and the synthetic occlusions we use in our experiments. Next, we provide more details for implementing VEN and the Mimic networks. We also provide further clarification regarding our evaluation metrics, and the evaluation protocol we use on the GREW dataset. We also evaluate VEN on the occlusion classification and the occlusion amount regression task, along with discussing the additional overhead introduced by VEN. Further, we provide more information regarding the reproducibility of our experiments, and we also evaluate our method on multiple different occlusion settings. Next, we provide some results on indoor gait recognition datasets like CASIA-B and OUMVLP.
Lastly, we discuss some failure cases of our model. 

\section{BRIAR Dataset}
BRIAR\cite{Cornett_2023_WACV} is a dataset collected for Person Re-ID in challenging outdoor conditions. It comprises of both images and video modalities, however, we utilize only the videos for this work. Some videos contain only the face information, which we discard while evaluation our gait recognition approach.

The subset of BRIAR we utilize in our experiments comprises approximately 60,000 videos for training and 10,000 videos for testing, each with a duration ranging from 1 to 2 minutes, recorded at 30 frames per second (fps). This extensive collection of videos offers a diverse array of gait sequences captured under various conditions, enabling comprehensive training and evaluation of gait recognition models.

The dataset encompasses a wide range of viewpoints and distances, including indoor controlled environments, close-range elevated viewpoints, and aerial perspectives captured by Unmanned Aerial Vehicles (UAVs). Distances from the subjects to the cameras span from 100 meters to 1000 meters, introducing varying levels of spatial resolution and turbulence challenges associated with long-distance capture. Furthermore, the use of UAVs with moving cameras adds another layer of complexity to the dataset.

\begin{figure*}
\centering
    \resizebox{0.8\linewidth}{!}{
        \centering
        \includegraphics{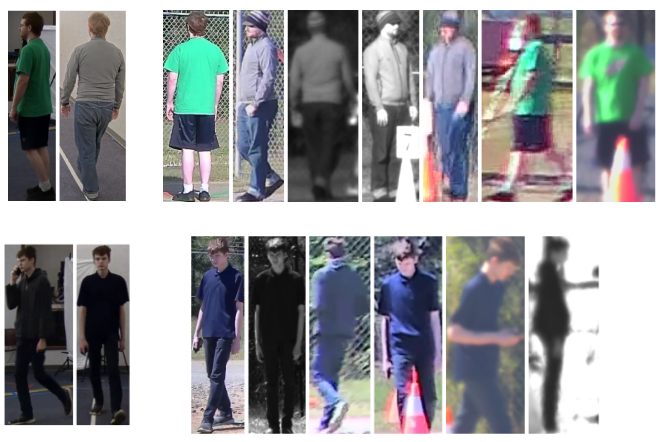}
    }
    \caption{Some more sample frames taken from videos present in the BRIAR dataset. Subjects have consented to use of these images. Each row consists of images of one subject. The two leftmost images in both rows show examples of the indoor, controlled gallery sequences. The remaining images are captured outdoors, and they make up the probe set. As the distance increases from left to right, the quality of the frames drops significantly.}
    \label{fig:more-briar-examples}
\end{figure*}

Each distance category in the dataset employs different sensors, capturing gait sequences in both black-and-white and color formats. Even within the color spectrum, different cameras introduce variations in image quality, contrast, color balance, dynamic ranges, and lens distortions. Some of these variations can be seen in \cref{fig:briar-examples} of the main paper. In \cref{fig:more-briar-examples}, we visualize some more frames captured from the BRIAR dataset, including some examples from the controlled indoor sequences. This huge diversity in the image quality necessitates robustness in gait recognition models to a large number of such variations.

The dataset encompasses a wide range of walking conditions, including random walks, structured walks, carrying a large cardboard box, wearing backpacks, using cellphones while walking, and even scenarios where subjects point at cameras while walking. These diverse conditions introduce variations in gait patterns, postures, and object interactions, enhancing the dataset's realism and applicability to real-world scenarios.

BRIAR also provides a predefined evaluation protocol along with a probe-gallery split for assessing gait recognition performance. Indoor sequences captured in controlled environments serve as the gallery set, while outdoor sequences, presenting more challenging scenarios, are designated as probe samples. Notably, standing sequences are excluded from the evaluation protocol to focus on walking-based gait recognition.

Different sets of videos of the same subject in different clothing are also collected. Further, some videos have significant occlusions present where the lower portion of the subject is not visible. With these large variations in acquisition conditions, atmospheric turbulence and changes in illumination introduced by long ranges and different camera sensors, the outdoor portion of the dataset is the more challenging part and is meant to be used as the probe set. 

In the indoor setups the subjects have different clothing conditions and either a random or structured way of walking. However, there are many more viewpoint variations in these indoor environments. But the indoor data has much smaller variations in illumination, turbulence, and noise compared to the outdoor dataset, and is thus meant to be used as the gallery set.

Only subjects who have explicitly consented to appear in the videos are included in the dataset, ensuring compliance with ethical guidelines and data protection regulations. Furthermore, subjects are given the option to decide whether their images can be used for publication purposes, with only those consenting to both inclusion in the dataset and publication being featured in visualizations within the paper.

\section{Synthetic Occlusions}

\paragraph{\textbf{Dynamic Occlusions:}}
To simulate dynamic occlusions, we place moving patches of various sizes in the video. Some examples have been shown in \cref{fig:dynamic-consistent} of the main paper. These occlusions try to simulate small stationary/moving obstacles which might obstruct the subject from camera view as the subject moves. Stationary objects like tall grass, trees, traffic signs and poles can block the subject, but as the camera follows the subject, the occlusion pattern from these objects appears to be moving from the frame of reference of the subject. 

\begin{figure}
    \centering
    \includegraphics[width=\linewidth]{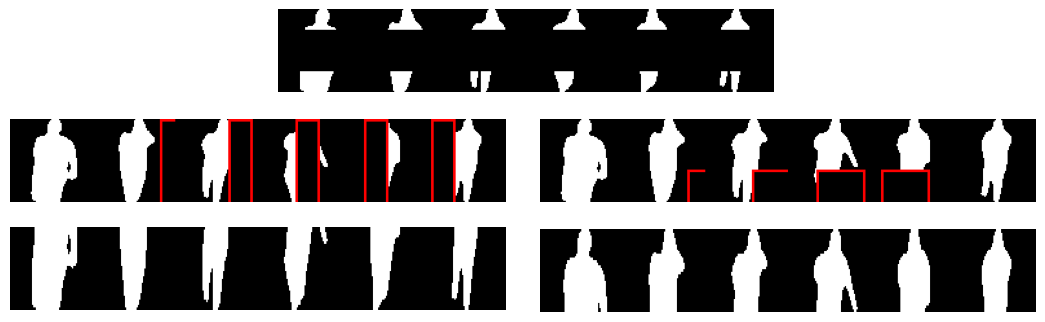}
    \caption{\label{fig:more-occ-vis} More visualizations of the synthetic occlusions on a video sequence taken from the GREW dataset. The top row shows middle occlusions. The second tow shows dynamic occlusions, and the bottom row shows top and bottom occlusions. The occlusion patch is shown with a red boundary in dynamic occlusions for visualization purposes only. In the synthetic dynamic occlusions, the width of the patch is more when the patch does not cover the entire height of the video. If it covers the whole height of the frame, the patch is relatively thinner. In top and bottom occlusions, the occluded portion is cropped out and the remaining frame is resized to the original height and width.
    }
    
\end{figure}

We consider two types of patches - small rectangular patches which can not cover the entire height of the frame, or tall rectangular patches which cover the entire height of the frame. The occlusion type to be applied on the video is randomly chosen to be either a small rectangular patch, or a tall rectangular patch as shown in \cref{fig:more-occ-vis}. If it is a small rectangular patch, the height and width of the patch are randomly chosen from the range $R$, which is set to (0.4, 0.6). If a tall rectangular patch is to be applied, the width of the patch is chosen within a different range $R_t$. $R_t$ is a smaller range than $R$ because we assume that tall objects like poles are thin and may not cover the entire width of the frame. We set $R_t$ to be $(0.2, 0.4)$, meaning the width of the tall patch may be between 20\%-40\% of the width of the frame. 

Since these patches move acorss the frame, the direction and speed of movement needs to be decided. The direction of motion is randomly chosen to be from left to right or right to left.
For the speed of movement, we visualize patches with different speeds and empirically decide which range of patch speeds look the most realistic for dynamic occlusion. We set the range of speeds of these moving patches $R_s = (0.5, 1.0)$ pixels/frame. Thus, for each video, the speed of the patch is also selected randomly within this range.    

Some more examples of the synthetic dynamic occlusions we use are shown in \cref{fig:more-occ-vis} of this supplementary material section.

\paragraph{\textbf{Consistent Occlusions:}}

We use three types of consistent occlusions, namely 1) top occlusion, where the torso and head of the person may be occluded; 2) bottom occlusion, where the legs and lower body may be occluded, and 3) middle occlusions, where the middle part of the body is occluded. The portion of the frame to be cropped out is chosen randomly from the fixed range $R$. 

In dynamic and middle occlusions, the occlusion patch zeros out the pixel values of the occluded region. However, in the case of top and bottom occlusions, the occluded portion of the frame is cropped out completely. The remaining part of the frame is then resized to the fixed $H \times W = 64 \times 64$ size of the original frame. Since we work with binary silhouette masks, resizing using linear interpolation causes the output image to become 8-bit non-binary integer type. Thus, we re-binarize the resized image using a threshold of 128. This simulates how an object detector would detect a subject in the case of real consistent occlusions. Some more examples of the consistent occlusions we use are shown in \cref{fig:more-occ-vis}.

\section{Visibility Estimation Network}
VEN, $\mathcal{V}$, is a three layer convolutional neural network with one hidden linear layer and two parallel linear heads for the classification and regression tasks. The complete architecture of $\mathcal{V}$ is presented in \cref{tab:ven-arch}.

\begin{table}[h]
\centering
\resizebox{0.6\linewidth}{!}{%
\begin{tabular}{c|c|c}
\textbf{Layer Name} & \textbf{Input shape} & \textbf{Output Shape}  \\ \hline
Conv1               & 64 $\times$ 64 $\times$ 1          & 64 $\times$ 64 $\times$ 32           \\
ReLU, MaxPool1      & 64 $\times$ 64 $\times$ 32         & 32 $\times$ 32 $\times$ 32           \\
Conv2               & 32 $\times$ 32 $\times$ 32         & 32 $\times$ 32 $\times$ 64           \\
ReLU, MaxPool2      & 32 $\times$ 32 $\times$ 64         & 16 $\times$ 16 $\times$ 64           \\
Conv3               & 16 $\times$ 16 $\times$ 64         & 16 $\times$ 16 $\times$ 128          \\
ReLU, MaxPool3      & 16 $\times$ 16 $\times$ 128        & 8 $\times$ 8 $\times$ 128            \\
AdaptiveAvgPool     & 8 $\times$ 8 $\times$ 128          & 128                    \\ \hline
FC1                 & 128                  & 64                     \\
Classification Head                & 64                   & 3      \\
Regression Head     & 64            & 1
\end{tabular}
}

\caption{\label{tab:ven-arch} The architecture of VEN. It is a three layer convolutional neural network, with one hidden linear layer, and two parallel linear heads which can predict the type and amount of occlusion in the input.}
\end{table}

Based on how many occlusion types the mimic network is supposed to be trained on, the classification head classifies the input into the occlusion classes or the no occlusion category. The cross entropy loss is used to train VEN through the classification head, so the network gains occlusion type awareness.

The above mentioned classes are simply broad categories of occlusion types, and the amount of synthetic occlusion within one category can also vary within a range $R$. In our experiments, we set the range of occlusions $R$ to be $(0.4, 0.6)$, so the output of the regression head is trained to be close to 0 when there are no occlusions in the input, and $x$ when the amount of occlusion is $x$. $x$ is sampled uniformly within the range $R$ for each video. As seen from \cref{tab:ablation-mimic-heads} of the main paper, the regression task helps the network gain occlusion amount awareness as well.

\subsection{Additional overhead of VEN}
VEN introduces additional parameters to the backbone during the inference stage - in terms of an extra convolutional network to generate occlusion relevant features. 
Specifically, the architecture of VEN we use introduces 0.1M additional parameters at inference time. This is relatively small compared to the gait recognition backbone - for example, GaitBase has roughly 7M parameters. 

The number of additional parameters introduced is exactly the same as the occlusion detector in \cite{occ_aware}, since the only difference between VEN and \cite{occ_aware} is the occlusion regression head which is discarded during inference.
It should be noted that we are counting only the parameters used during the inference stage and not in the training stage. This means we are excluding the BNNeck layer in GaitBase, and the classification and regression heads in VEN in the numbers reported above. These layers are used only during training time. 

\section{Implementation Details}

In this section, we elaborate on the implementation details of our method. For ease of reproducibility, we have released the source code.

\paragraph{\textbf{Preprocessing: }}
If the input video $S^i$ is of RGB modality, we first extract binary masks from the video. For this, we use Detectron2 \cite{wu2019detectron2}, to obtain the masks around the subject for each frame. This is done to filter out any covariates like background, color and texture hampering gait recognition.

On obtaining the silhouette masks, we center the subject and resize all the frames to a uniform size of $H \times W$ similar to \cite{opengait}. In the frames where no subject is detected, we leave an empty black frame in the output video to keep the number of frames consistent.

\label{sec:supp-imple}
\paragraph{\textbf{VEN:}}
We train VEN using the Adam optimizer \cite{KingmaB14adam} with a learning rate of 1e-4. During training, the classification loss $L_{ce}$ and the regression loss $L_r$ are multiplied by loss weights $\lambda_{ce}$ and $\lambda_{r}$ to calculate the final loss $L$ for the backward pass as shown below

\begin{equation}
    L = \lambda_{ce}L_{ce} + \lambda_rL_r 
\end{equation}

Empirically, we find that setting $\lambda_{ce} = 1.0$ and $\lambda_r = 10.0$ yields the best performance in the proxy tasks of occlusion classification and regression.

\paragraph{\textbf{Mimic Network:}}
For pretraining of $\mathcal{F}_t$, Triplet and Cross Entropy losses are used as done by \cite{opengait}. In the distillation stage, the multi-instance correlational distillation loss is modelled as a TripletMarginLoss\cite{balntas2016triplet} with a margin $m = 0.05$. 

For training GaitGL models, we randomly sample $n = 30$ contiguous frames from the full video. For GaitBase and DeepGaitV2, $n$ is chosen uniformly between (20, 40) for each video. A batch size of (32, 4) is used for training, meaning that each batch of training data has 32 identities and 4 sequences per identity. 

Apart from the randomly generated synthetic occlusions, we also use data augmentation techniques like Random Horizontal Flipping, Random Cropping and Random Perspective to train $\mathcal{F}_t$ and $\mathcal{F}_m$. We use the same data augmentation settings as used by \cite{opengait}.

For introducing occlusion-relevant features into the gait recognition backbone, we adopt the approach used by \cite{occ_aware}. Specifically, we utilize the fully connected layers in the later parts of the gait recognition backbones as positions for inserting the occlusion features provided by VEN.

\section{Evaluation Details}
We use the Top-K rank retrieval accuracy to evaluate gait recognition performance for all datasets. In addition, we perform verification as well on the BRIAR\cite{Cornett_2023_WACV} dataset, computing the True Acceptance Rate(TAR) at a False Accept Rate(FAR) of 0.01. 

The respective datasets provide their protocols which tell us which video sequences should be used as probes and which ones should be used as a part of the gallery set. As described in \cref{sec:grew-eval-protocol}, we use a local evaluation protocol for GREW introduced by \cite{opengait} so that we can evaluate our models locally.

To evaluate our approach on synthetic occlusions, we use the same gallery-probe split from the dataset protocol but introduce synthetic occlusions in each video during the data loading stage. 
It should be noted that the occlusion type and amount is chosen for each video independently - thus, there is no correlation between the occlusions in the probe and gallery sequences of a particular subject. The probe and gallery sequences may have different types and amounts of occlusions, which makes our task formulation more challenging and better suited for practical application.

We compute the gait signatures for each element in the gallery set, and compare each probe signature with each gallery to find the Top-K subject matches. We use Euclidean distance to compare probe and gallery elements. If the true identity of the probe is present in the Top-K matches, the probe is considered to be recognized correctly. The percentage of probes recognized correctly is reported as the rank retrieval accuracy for Rank-K. This process is repeated for different values of K for more comprehensive evaluation.

\section{GREW evaluation protocol}
\label{sec:grew-eval-protocol}
The GREW dataset\cite{zhu2021gait} does not provide identity labels for their probe set. Hence, local evaluation of a model is not possible. According to the official protocol, the matching scores for each gallery and probe video have to be uploaded on the GREW competition website, which computes the accuracy of the model. This is limiting for our experiments, especially since we can not compare our results to other papers directly and have to re-train previous methods on our synthetically occluded data.

Thus, as mentioned in the main paper, we use a slightly different evaluation protocol for GREW which enables local evaluation. This protocol was introduced by \cite{opengait} and has also been used in some existing works\cite{occ_aware}. We explain this modified protocol below.

Each of the 6,000 subjects in the test set of GREW have two gait sequences, giving a total of 12,000 videos in the test set.  Instead, one video of each subject is chosen as gallery and the other video is chosen as the probe, giving each subject one sequence in the gallery set. The rank-retrieval task is performed on this probe-gallery split and corresponding Rank-K metrics are computed. For the purposes of this protocol, the unlabelled videos in the `probe' directory of GREW are ignored.

As a sanity check to confirm whether we reproduce the existing methods correctly, we take the teacher model $\mathcal{F}_t$, which is trained on complete videos, and evaluate it directly on the original GREW data without any occlusions. We evaluate this model both on the official GREW protocol and the modified protocol. The official protocol results are directly comparable to the results in the original papers, indicating we have correctly reproduced these works.
The results are summarized in \cref{tab:local-vs-website-eval}. 
We observe that the local evaluation protocol consistently has lower rank retrieval accuracy than when we use the official protocol. We think this is because in our local evaluation protocol, there is only one gallery sequence per subject. However, in the official protocol, there are two gallery sequences per subject - making it a bit easier for the model to match the probe to the appropriate gallery.

\begin{table}
\centering
\begin{tabular}{c|c|c}
Rank-1/Rank-5 & Local Evaluation & Official protocol  \\ 
\hline
GaitBase\cite{opengait}    & 55.3/72.1        & 59.9/74.7           \\
DeepGaitV2\cite{deepgaitv2}  & 73.1/85.3        & 78.4/88.6          
\end{tabular}
\caption{\label{tab:local-vs-website-eval}
Comparing the evaluation results on the official protocol (using the submission website) with our local evaluation protocol. The models are trained and evaluated on the original GREW dataset, making them identical to the teacher model $\mathcal{F}_t$ in our method. The numbers are the Rank-1/Rank-5 accuracies on the GREW dataset. The numbers in the `official protocol' column are directly comparable to the results in the corresponding papers. We perform this experiment as a sanity check to see whether we reproduce the original methods correctly.
}
\end{table}

\section{VEN Results}
We train VEN on the proxy tasks of occlusion type classification and occlusion amount regression. This helps VEN to learn occlusion-relevant features, which are useful for occluded gait recognition as seen in \cref{tab:ablation-mimic-heads} - when we remove VEN, the `vanilla' mimic network performs worse.

However, we also perform a sanity check evaluation of VEN on these same proxy tasks themselves, similar to \cite{occ_aware}. More specifically, we compute the classification accuracy of the occlusion type, and the mean squared error in the regression amount for frames taken from the test set, of the same or a different dataset than it was trained on. This helps us get an idea of the robustness of VEN to domain shifts.

The results of this evaluation of VEN are shown in \cref{tab:ven-eval}. We observe that VEN is able to perform well on these proxy tasks and is also able to generalize to other domains/datasets.

\begin{table}
\centering
\resizebox{0.8\linewidth}{!}{%
\begin{tblr}{
  cells = {c},
  cell{1}{1} = {c=2,r=2}{},
  cell{1}{3} = {c=2}{},
  cell{3}{1} = {r=2}{},
  vline{2} = {1}{},
  vline{4} = {2}{},
  vline{2-4} = {3}{},
  vline{2-4} = {4}{},
  vline{3-4} = {1}{},
  vline{3-4} = {2}{},
  hline{2} = {3-4}{},
  hline{3} = {-}{},
  hline{4} = {2-4}{},
}
Accuracy/MSE &       & Train        &              \\
             &       & BRIAR        & GREW         \\
Test         & BRIAR & 99.9/0.00060 & 99.7/0.00060 \\
             & GREW  & 99.0/0.00161 & 99.9/0.00031 
\end{tblr}
}
\caption{\label{tab:ven-eval}Cross dataset evaluation of VEN, on BRIAR and GREW datasets on top, bottom and no occlusion classes. The proxy tasks of occlusion type classification and occlusion amount regression are used to perform this evaluation. The first value of each cell represents classification accuracy of the occlusion type, the second value represents the MSE in occlusion amount prediction. VEN performs reasonably well on both these tasks, indicating that it has learned occlusion-relevant information. We can also see that VEN is robust to domain shifts, since switching to a different dataset for evaluation does not decrease the performance as much.}
\end{table}

\section{Additional Experiments}

\subsection{Reproducibility of results}
Since our evaluations are based on random occlusions, we also perform multiple evaluations of our network to check whether the results are reproducible across evaluation runs. 
Hence, we perform 10 repeat evaluation runs on the mimic network on the GREW dataset using the GaitBase backbone. We observe a standard deviation of 0.35\% in the Rank-1 accuracy, indicating there is very little change in overall performance due to the introduction of randomness in the evaluation process.

\subsection{Changing occlusions within a video}
In all our previous experiments, we have assumed that the occlusion type remains the same across the video. Here, we conduct an experiment where the occlusion type can change among the frames in the video. More specifically, we flip the occlusion type from top to bottom and vice versa in the middle of a video to see whether this hampers the performance of the model. Indeed, when comparing it to \cref{tab:main-results-mimic} of the main paper, there is a drop in performance for all methods. However, we observe that the mimic network still outperforms other methods in this changing occlusion scenario as shown in \cref{tab:change-occ-middle}. This further demonstrates the generalizability of our method.

\begin{table}
\centering

\begin{tabular}{c|cc}
 Changing occlusion types  & Rank-1 & Rank-5  \\ 
\hline
Baseline-2    & 9.73   & 19.52   \\
Occlusion Aware\cite{occ_aware}     & 14.63  & 27.65   \\
Mimic Network & 16.05  & 29.75  
\end{tabular}

\caption{\label{tab:change-occ-middle}
Effect of flipping the occlusion type in the middle of a video, between top and bottom occlusion cases. The mimic network is able to deal with changing occlusion types better than other methods even though it too has not seen such data during training. This further demonstrates the generalizability of our method.
}
\end{table}

\subsection{Difficulty of different occlusion types}
\label{sec:difficulty-diff-occs}
In this section, we compare the relative difficulty of different occlusion types for the mimic network. For this, we take a model trained on top, bottom and middle occlusions. However, during evaluation, we restrict the occlusions to one type at a time, for top, bottom and middle occlusions.
The results are presented in \cref{tab:one-occ-type}. As one might expect, bottom occlusions are the most difficult for the network to work on, since the legs - the body part where the most obvious gait patterns appear - are not visible in these occlusions. Interestingly, top and middle occlusions are roughly equally difficult for the model. 

\begin{table}
\centering
\begin{tabular}{c|cc}
Restricting Occlusion Types & Rank-1 & Rank-5  \\
\hline
Top occlusion only        & 30.17  & 47.27   \\
Middle occlusion only     & 29.62  & 48.67   \\
Bottom occlusion only     & 14.17  & 26.6   
\end{tabular}
\caption{\label{tab:one-occ-type}
Restricting the occlusion types during evaluation of the mimic network. This shows the relative difficulty of different occlusion types. We can see that bottom occlusions are the most difficult for the mimic network, since the gait is much more difficult to observe when legs are not visible.
}
\end{table}

\subsection{Training on all occlusion types}

So far, our experiments have focused on top and bottom occlusions, with some analysis being done on middle and dynamic occlusions. 
Since it is not practical to train on all possible occlusion types that may occur, a model needs to be able to generalize to newer occlusion types. Hence, we do a generalizability evaluation in the `Zero-shot evaluation' columns of \cref{tab:new-occs} in the main paper, taking a model trained on top and bottom occlusions and applying it on middle or dynamic occlusions. 

However, if a particular type of occlusion is anticipated in the task setup - maybe due to camera placement constraints in the final use case - it is possible to prepare the model for the specific occlusion type. Generally, the trend is that training a model on a specific occlusion type improves performance on that occlusion set. This is demonstrated in the Training columns of \cref{tab:new-occs} of the main paper. In these training columns, we add an additional occlusion type (middle or dynamic) in the training set, in addition to top and bottom occlusions. 

While training on newer occlusion types helps, we re-iterate that it is not practical to train (and evaluate) a gait recognition model for every possible occlusion type which might occur. 
Hence, in our main paper, we focus on only a limited set of occlusion types (top and bottom occlusions) for training the model in our main results and then try to see whether this model generalizes well to other occlusion types - rather than training the model on all the occlusion types at once. 

In this supplementary material section, 
for completeness, we also present the results from training new networks on the combined set of all occlusion types we have used in our experiments - Top occlusions, Middle occlusions, Bottom occlusions, dynamic small patch occlusions and dynamic tall patch occlusions. The results are summarized in \cref{tab:occ-5-types}. We observe that the mimic network is still able to outperform other approaches on this combined occlusion set.

\begin{table}
\centering
\begin{tabular}{c|cc}
    All occlusion types            & Rank-1 & Rank-5  \\ 
\hline
Baseline-2      & 30.3  & 46.2   \\
Occlusion Aware & 35.5   & 52.4    \\
Mimic Network   & 43.0    & 60.2  
\end{tabular}
\vspace{1mm}
\caption{\label{tab:occ-5-types}
Performance of different methods on the mixed occlusion set, comprising of top, middle, bottom, dynamic small and dynamic tall patches. These results are on the GREW dataset using the GaitBase backbone.
}
\end{table}

\subsection{Different occlusion ranges}

\begin{table}
\centering
\resizebox{0.7\linewidth}{!}{%
\begin{tabular}{c|cc}
Range   & Rank-1       & Rank-5        \\ 
\hline
40-60\% & 28.38 (0.51) & 45.43 (0.63)  \\
30-50\% & 33.62 (0.61) & 51.33 (0.71)  \\
20-40\% & 40.25 (0.73) & 57.68 (0.80)  \\
10-30\% & 45.35 (0.82) & 61.85 (0.86) 
\end{tabular}
}
\caption{\label{tab:expt-occ-ranges}
Performance of the mimic network on multiple ranges of top and bottom occlusion, on the GREW dataset using GaitBase backbone. The RP values are shown in (.). The first row denotes the experiments with the standard occlusion range we use throughout the paper. Both the rank retrieval accuracy and the RP increase as we reduce the occlusion range, as expected.
}
\end{table}

In our work, we pick an occlusion range $R$, which we set to (0.4, 0.6) for most of our experiments in consistent occlusions. This means the amount of occlusion is randomly sampled to be 40-60\% of the frame dimension. In this section, we explore how the performance changes on different occlusion ranges. We present the results in \cref{tab:expt-occ-ranges}. As expected, the performance improves as we reduce the amount of occlusion. This is reflected in both the rank retrieval accuracy and the RP values.

\paragraph{}
Our results on different occlusion types or different occlusion settings are some basic analyses which we performed to investigate how different occlusions impact the gait recognition problem. This is in no way a comprehensive analysis of which occlusion setting is more practical, or more likely to occur in real scenarios. 
We maintain that the focus of this work is to propose an approach which works better than other approaches for some given occlusion settings.   
The analysis of different occlusion types is out of the scope of this work and we leave it up to future research.

\subsection{Speed of adaptability}
In the adaptability scenario, we extend the capability of training a model on a new type of occlusion in addition to the old occlusions it was trained on. 
We discuss how effectively the networks are able to adapt to different occlusion types in \cref{tab:new-occs}.
Here, we discuss the speed of this adaptation - how much re-training is required to adapt the model to a new occlusion type. We discuss the GaitBase backbone on the GREW dataset.

GaitBase is trained on 180,000 iterations on the GREW dataset, according to the settings used by \cite{opengait}. We also train the model on top and bottom occlusions for the same number of iterations. However, when adapting the model to a new occlusion type, say middle occlusions, we train the model further until the training loss converges with the new occlusion set. We observe that the training loss converges at around 20,000 iterations. Thus, the network is able to adapt to a new occlusion set with additional training of 20,000 iterations, or roughly 11\% of additional training time.

\subsection{Results on indoor datasets}
Our main focus is on the outdoor, in-the-wild scenarios where occlusion is more likely to occur. Hence, we focus our experiments on in-the-wild datasets like GREW, Gait3D and BRIAR. 
However, for completeness, we evaluate our approach on indoor datasets using the simulated top and bottom occlusions as well. We report the Rank-1 accuracy on Normal Walking (NM), Baggage (BG) and Cloth Changing (CL) for the CASIA-B dataset according to the standard protocol \cite{opengait}. 
We also report the Rank-1 accuracy on NM for OUMVLP according to the standard protocol used in \cite{opengait}. 
Additionally, we report the RP values for all these metrics in (.). The results are presented in \cref{tab:indoor-results}. They show a similar trend, where the mimic network is able to outperform other methods on occluded gait recognition.

\begin{table}
\centering
\resizebox{\linewidth}{!}{%
\begin{tabular}{c|ccc|c}
\multirow{2}{*}{Method}       & \multicolumn{3}{c|}{CASIA-B}                                          & OUMVLP     
\\
\cline{2-5}
                              & NM                    & BG                    & CL                    & NM                     \\ 
\hline
Baseline-1                    & 22.76 (0.23)          & 20.45 (0.22)          & 10.03 (0.13)          & 2.09 (0.02)            \\
Baseline-2                    & 31.35 (0.32)          & 24.56 (0.26)          & 13.98 (0.18)          & 14.47 (0.16)           \\
Occlusion Aware               & 52.80 (0.54)          & 47.10 (0.50)          & 33.46 (0.43)          & 17.65 (0.19)           \\ 
\hline
\textbf{Mimic Network (ours)} & \textbf{69.42 (0.71)} & \textbf{57.12 (0.61)} & \textbf{37.35 (0.48)} & \textbf{25.42 (0.28)} 
\end{tabular}
}
\caption{\label{tab:indoor-results}
Rank-1 accuracy for different conditions on CASIA-B and OUMVLP datasets, on top and bottom occlusions. The corresponding RP values are reported in (.). The mimic network is able to outperform other approaches on the indoor datasets as well.
}
\end{table}

\section{Failure case Analysis}

Though our model is able to improve performance on occluded gait recognition, it is not perfect. We have discussed the relative difficulty of different occlusion types in \cref{sec:difficulty-diff-occs}, concluding that the model is more likely to fail in bottom occlusions. 
From \cref{tab:expt-occ-ranges}, we also empirically confirm that the model is more likely to fail when the occlusion is more severe.    

\begin{figure}
    \centering
    \includegraphics[width=0.95\linewidth]{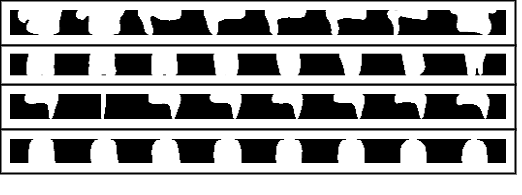}
    \caption{\label{fig:failure-cases}
    Visualization of some of the failure cases of the model - the misclassified probes on the GREW dataset. We see that in a lot of these examples, it is not even clear which body part is being shown; it is extremely difficult to identify these subjects clearly because of the lack of discriminative information in the input.
    }
\end{figure}

In this section, we discuss some of the specific failure cases of the model. We have shown some examples of probes taken from the GREW dataset which are misidentified by our model in \cref{fig:failure-cases}. 
In some of these probes, it is not possible to make out which body part is present in the input. This makes it difficult to extract its correlations with the missing body parts.
In some probes, only the head is visible and there is barely any motion present in the input. It becomes difficult to recognize the subject from a generic circular silhouette without any characteristic motion.

From this observation, we can conclude that the model can not perform on any input - some basic body parts have to be recognizable for the model to extract correlations with the other body parts. 
Further, temporal information is necessary for the model to recognize the gait of the subject. Without any motion in the input, the model will likely not be able to identify the subject correctly.
These are recognized as potential situations where there is scope for improvement and we leave this to future work. 

This error analysis has been performed on the GREW dataset with the mimic network using the GaitBase backbone. 

\section{Cross-Entropy loss}
In our experiments in \cref{tab:ablation-kd}, we observe that adding cross-entropy loss to our Multi-instance Correlational KD (MiCKD) loss for training the mimic network ends up reducing performance. Here, we try to analyse this issue.

\begin{figure}
    \centering
    \includegraphics[width=\linewidth]{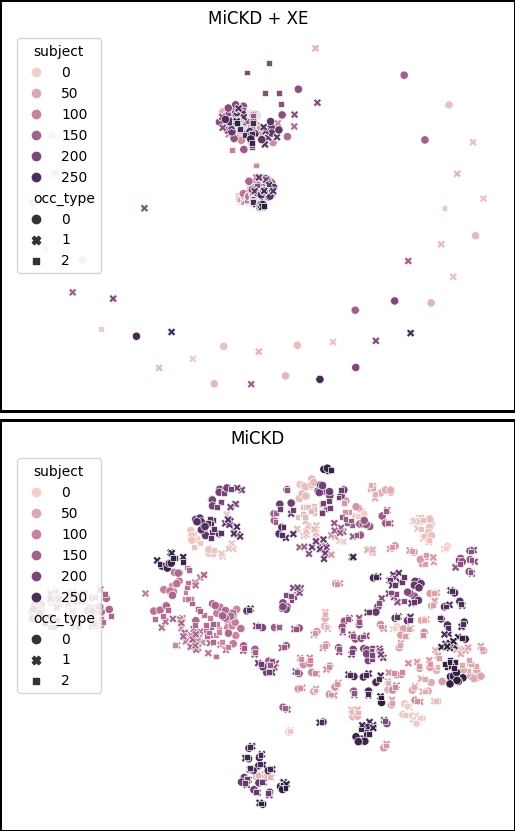}
    \caption{Visualization of t-SNE features with and without using the cross entropy (XE) loss along with the proposed MiCKD loss. Different colors denote different subjects and different shapes denote different occlusion types. Features are relatively better clustered when cross entropy loss is not used. }
    \label{fig:mickd-xe-tsne}
\end{figure}

We follow the cross-entropy formulation used in \cite{opengait}, where we use a BNNeck layer on the embeddings before applying cross entropy loss. This is shown to stabilize training and yield better performance. 
However, this cross-entropy (XE) formulation hurts the performance of the model when used along with the MiCKD loss. To investigate this further, we plot the t-SNE features of some samples from the GREW dataset in \cref{fig:mickd-xe-tsne}. 

We observe an interesting pattern in the MiCKD + XE embeddings, where a lot of the embeddings are clustered in the center while some are scattered further away. The proximity of most of the embeddings in the center makes it difficult to match probe embeddings to the proper gallery. 
On the other hand, the embeddings in the MiCKD figure are clustered better than MiCKD + XE. Based on this observation, we conclude that when MiCKD and XE are used together, the embeddings are not able to cluster well - in other words, MiCKD and XE loss do not go well together. 

Based on the results of \cite{opengait}, we know that Triplet Loss and cross-entropy loss go well together. Thus, one might try to replace MiCKD with the Triplet Loss used in \cite{opengait}, and use it along with cross entropy loss. 
However, this formulation when applied to occluded data is the same as Baseline-2 in \cref{tab:ablation-kd}; it performs worse than even MiCKD + XE. It should be noted that VEN has been removed in these set of experiments, and we are dealing with a vanilla variant of the mimic network, one without the VEN.

\end{document}